\documentclass[letterpaper,twocolumn,10pt]{article}
\usepackage{usenix,epsfig,endnotes}

\usepackage{bm}

\usepackage{mathtools}
\usepackage{amsfonts}
\usepackage{algorithm}
\usepackage[noend]{algpseudocode}
\usepackage{babel,blindtext}

\usepackage{enumitem}

\usepackage{float}

\usepackage{multirow}
\usepackage{tabularx}

\usepackage{subfigure}
\usepackage{amsthm}
\usepackage{algorithmicx}

\usepackage{pgfplots}
\pgfplotsset{compat=1.9}
\usepackage{balance}
\usepackage{flushend}

\newtheorem{thm}{Theorem}
\newtheorem{lem}{Lemma}
\newtheorem{prop}{Proposition}
\newtheorem{cor}{Corollary}

\newtheorem{mydef}{Definition}

\newcommand{\norm}[1]{||#1||}

\newcommand{\bigO}{\mathcal{O}}
\newcommand{\domainD}{\mathcal{D}}

\newcommand{\domainP}{\mathcal{P}}
\newcommand{\seta}[1]{\lbrace #1 \rbrace}

\newcommand{\bA}{\mathbf{A}}
\newcommand{\bD}{\mathbf{D}}

\newcommand{\ba}{\mathbf{a}}
\newcommand{\bc}{\mathbf{c}}
\newcommand{\bd}{\mathbf{d}}
\newcommand{\be}{\mathbf{e}}
\newcommand{\bmf}{\mathbf{f}}

\newcommand{\br}{\mathbf{r}}
\newcommand{\bs}{\mathbf{s}}
\newcommand{\bv}{\mathbf{v}}
\newcommand{\bx}{\mathbf{x}}
\newcommand{\by}{\mathbf{y}}

\newcommand{\rhigh}[1]{r^{(#1)}}
\newcommand{\shigh}[1]{s^{(#1)}}
\newcommand{\vhigh}[1]{v^{(#1)}}
\newcommand{\xhigh}[1]{x^{(#1)}}

\begin{document}

\date{}

\title{\Large \bf Efficient Optimization of Dominant Set Clustering with Frank-Wolfe Algorithms}

\author{
{\rm Carl Johnell}\\
Department of Computer Science and Engineering\\
Chalmers University of Technology\\
Email: {cjohnell@gmail.com}
\and
{\rm Morteza Haghir Chehreghani}\\
Department of Computer Science and Engineering\\
Chalmers University of Technology\\
Email: {morteza.chehreghani@chalmers.se}
}

\maketitle
\thispagestyle{empty}

\begin{abstract}
We study Frank-Wolfe algorithms -- standard, pairwise, and away-steps -- for efficient optimization of Dominant Set Clustering. We present a unified and computationally efficient framework to employ the different variants of Frank-Wolfe methods, and we investigate its effectiveness via several experimental studies. In addition, we provide explicit convergence rates for the algorithms in terms of the so-called Frank-Wolfe gap. The theoretical analysis has been specialized to Dominant Set Clustering and covers consistently the different variants.

\end{abstract}


\section{Introduction}
Clustering plays an important role in unsupervised learning and exploratory data analytics \cite{jain2010}. It is used in applications from different domains such as network analysis, image segmentation, document and text processing, community detection and bioinformatics.
Given a set of $n$ objects with indices $V = \{1,...,n\}$  and the nonnegative pairwise similarities $\mathbf A = (a_{ij})$, i.e., graph $\mathcal G(V,\mathbf A)$ with vertices $V$ and edge weights $\mathbf A$, the goal is to partition the data into coherent groups that look dissimilar from each other. We assume zero self-similarities, i.e., $a_{ii}=0 \text{ } \forall i$.
Several clustering methods compute the clusters via minimizing a cost function. Examples are Ratio Cut \cite{ChanSZ94}, Normalized Cut \cite{Shi:2000:NCI}, 
Correlation Clustering \cite{Bansal02correlationclustering}, and shifted Min Cut \cite{Chehreghani17ICDM,abs-2110-13103}. For some of them, for example Normalized Cut, approximate solutions have been developed in the context of spectral analysis \cite{Shi:2000:NCI,Ng01onspectral}, Power Iteration Clustering (PIC) method \cite{LinC10PIC} and P-Spectral Clustering \cite{Buhler:2009PSpec,HeinB10PSpec}.  They are sometimes combined with greedy hierarchical clustering, e.g., as $K$-means is combined with agglomerative clustering \cite{ChehreghaniAC08}.

Another prominent clustering approach has been developed in the context of Dominant Set Clustering (DSC) and its connection to discrete-time dynamical systems and \emph{replicator dynamics} \cite{pavan2007,bulo2017}.
Unlike the methods based on cost function minimization, DSC  does not define a global cost function for the clusters. Instead, it applies the generic principles of clustering where each cluster should be coherent and well separated from the other clusters. These principles are formulated via the concepts of dominant sets \cite{pavan2007}. Then, several variants of the method have been proposed.
The method in  \cite{LiuLY13} proposes an iterative clustering algorithm in two Shrink and Expand steps. These steps are suitable for sparse data and lead to reducing the runtime of performing replicator dynamics. \cite{BuloTP09} develops an enumeration technique for different clusters via unstabilizing the underlying  equilibrium of replicator dynamics. \cite{pavan2003} proposes a hierarchical variant of DSC via regularization and shifting the off-diagonal elements of the similarity matrix.
\cite{chehreghani2016} analyzes adaptively the trajectories of replicator dynamics in order to discover suitable phase transitions that correspond to evolving fast clusters. Several studies demonstrate the effectiveness of DSC variants compared to other clustering methods, such as spectral methods \cite{pavan2007,LiuLY13,chehreghani2016,bulo2017}.

In this paper, we investigate efficient optimization for DSC based on Frank-Wolfe algorithms \cite{frank1956,lacoste2015,reddi2016} as an alternative to replicator dynamics.  Frank-Wolfe optimization has been successfully applied to several constrained optimization problems. We develop a unified and computationally efficient framework to employ the different variants of Frank-Wolfe algorithms for DSC, and we investigate its effectiveness via several experimental studies.  Our theoretical analysis is specialized to DSC, and we provide explicit convergence rates for the algorithms in terms of  Frank-Wolfe gap -- including pairwise Frank-Wolfe with nonconvex/nonconcave objective function for which we have not seen any theoretical analysis in prior work. In addition, we study multi-start Dominant Set Clustering that can be potentially useful for parallelizing the method.

We note that beyond DSC, replicator dynamics is used in several other domains such as evolutionary game theory \cite{Cressman10810}, theoretical biology \cite{PAGE200293}, dynamical systems \cite{phd_Cayuela},  online learning \cite{2164-6066_2020_4_365}, combinatorial optimization problems
\cite{Pelillo2001}, and several other tasks beyond clustering \cite{bulo2017}. Hence, our contribution opens novel possibilities to investigate more efficient alternatives in those paradigms.

\section{Dominant Set Clustering}
DSC follows an iterative procedure to compute the clusters: i) computes a dominant set using the similarity matrix $\bA$ of the available data, ii) peels off (removes) the clustered objects from the data, and iii) repeats until a predefined number of clusters have been obtained.\footnote{With some abuse of the notation, $n$, $V$, $\mathbf A$ and $\bx$ sometimes refer to the available (i.e., still unclustered) objects and the similarities between them. This is obvious from the context. The reason is that DSC performs a sequential procedure where at each step separates a cluster from the available unclustered data.}

Dominant sets correspond to local optima of the following quadratic problem \cite{pavan2007}, called standard quadratic problem (StQP).
\begin{align} \label{eq:stqp1}
\textrm{maximize } f(\bx) &= \bx^T\bA\bx \\
\textrm{subject to } \bx \in \Delta &= \left\lbrace \bx \in \mathbb{R}^n : \bx \geq \mathbf{0}^n \textrm{ and } \sum_{i=1}^n x_i = 1 \right\rbrace. \nonumber
\end{align}
The constraint $\Delta$ is called the standard simplex. We note that $\bA$ is generally not negative definite, and the objective function $f(\bx)$ is thus \emph{not} concave.

Every unclustered object $i$ corresponds to a component of the $n$-dimensional characteristic vector $\bx$. The support of local optimum $\bx^{*}$ specifies the objects that belong to the dominant set (cluster), i.e., $i$ is in the cluster if component $x^{*}_i >0$. In practice we use $x^{*}_i > \delta$, where $\delta$ is a small number called the cutoff parameter. Previous works employ replicator dynamics to solve StQP, where $\bx$ is updated according to the following dynamics.
\begin{equation}
\xhigh{t+1}_i = \xhigh{t}_i\frac{(\mathbf A \mathbf \bx_t)_i}{\bx_t^{\texttt{T}}\mathbf A \bx_t} \; , i=1,..,n\, ,
\label{eq:replicatorEq}
\end{equation}
where $\bx_t$ indicates the solution at iterate $t$, and $\xhigh{t}_i$ is the $i$-th component of $\bx_t$. We note the $\bigO(n^2)$ per-iteration time complexity due to the matrix multiplication.

In this paper we investigate an alternative optimization framework based on Frank-Wolfe methods.

\section{Unified Frank-Wolfe Optimization Methods}
Let $\domainP \subset \mathbb{R}^n$ be a finite set of points and $\domainD = \textrm{convex}(\domainP)$
its convex hull (convex polytope). The Frank-Wolfe algorithm, first introduced in \cite{frank1956}, aims at solving the following  constrained optimization.
\begin{equation}\label{eq:fw}
\max_{\bx \in \domainD} f(\bx),
\end{equation}
where $f$ is nonlinear and differentiable.
The formulation in \cite{lacoste2016} has extended the concavity assumption to arbitrary functions with  $L$-Lipschitz (`well-behaved') gradients.
Algorithm \ref{alg:fw:v1} outlines the steps of a Frank-Wolfe method to solve the optimization in \eqref{eq:fw}.

In this work, in addition to the standard FW (called FW), we also consider two other variants of FW: pairwise FW (PFW) and away-steps FW (AFW), adapted from \cite{lacoste2015}.  They differ in the way the ascent direction $\bd_t$ is computed.

\begin{algorithm}[htb]
    \caption{Frank-Wolfe pseudocode}
    \label{alg:fw:v1}
    \begin{algorithmic}[1] 
        \Procedure{PSEUDO-FW}{$f$, $\domainD$, $T$}
        \Comment{Function $f$, convex polytope $\domainD$, and iterations $T$.}
            \State Select $\bx_0 \in \domainD$
            \For{$t = 0, ..., T-1$}
                \State \textbf{if} $\bx_t$ is stationary \textbf{then break}
                \State Compute feasible ascent direction $\bd_t$ at $\bx_t$
                \State Compute step size $\gamma_t \in [0, 1]$ such that $f(\bx_t + \gamma_t\bd_t) > f(\bx_t)$
                \State $\bx_{t+1} := \bx_t + \gamma_t\bd_t$
            \EndFor
            \State \textbf{return} $\bx_t$
        \EndProcedure
    \end{algorithmic}
\end{algorithm}

From the definition of $\domainD$, any point $\bx_t \in \domainD$ can be written as a convex combination of the points in $\domainP$, i.e.,

\begin{equation} \label{eq:convexcomb}
    \bx_t = \sum_{\bv \in \domainP} \lambda_{\bv}^{(t)} \bv,
\end{equation}

where the coefficients $\lambda_{\bv}^{(t)} \in [0, 1]$ and $\sum_{\bv \in \domainP} \lambda_{\bv}^{(t)} = 1$. Define

\begin{equation} \label{eq:convexcoeffs}
S_t = \seta{\bv \in \domainP: \lambda_{\bv}^{(t)} > 0}
\end{equation}
as the set of points with nonzero coefficients at iterate $t$.
Moreover, let
\begin{align}
    \label{eq:fwstep1}
    \bs_t &\in \arg \max\limits_{\bs \in \domainD} \nabla f(\bx_t)^T\bs, \\
    \label{eq:fwstep2}
    \bv_t &\in \arg \min\limits_{\bv \in S_t} \nabla f(\bx_t)^T\bv.
\end{align}
Since $\domainD$ is a convex polytope, $\bs_t$ is the point that maximizes the linearization and $\bv_t$ is the point with nonzero coefficient that minimizes it over $S_t$.
Let $\bx_t$ be the estimated solution of \eqref{eq:fw} at iterate $t$ and define
\begin{align} \label{eq:fw:dir}
    \bd_t^{A} &= \bx_t - \bv_t, \nonumber \\
    \bd_t^{FW} &= \bs_t - \bx_t, \nonumber \\
    \bd_t^{AFW} &=  \left\{
	\begin{array}{ll}
	\bd_t^{FW}, & \textrm{if } \nabla f(\bx_t)^T\bd_t^{FW} \geq f(\bx_t)^T\bd_t^{A} \\
	\frac{\lambda_{\bv_t}^{(t)}}{1 - \lambda_{\bv_t}^{(t)}}\bd_t^{A},  & \textrm{otherwise}
	\end{array}
    \right. \nonumber \\
    \bd_t^{PFW} &= \bs_t - \bv_t
\end{align}
respectively as the away, FW, pairwise, and away/FW directions (to be maximized). The FW direction moves towards a `good' point, and the away direction moves away from a `bad' point. The pairwise direction shifts from a `bad' point to a `good' point \cite{lacoste2015}. The coefficient with $\bd_t^{A}$ in $\bd_t^{AFW}$ ensures the next iterate remains feasible.

An issue with standard FW, which PFW and AFW aim to fix, is the zig-zagging phenomenon. This occurs when the optimal solution of \eqref{eq:fw} lies on the boundary of the domain. Then the iterates start to zig-zag between the points, which negatively affects the convergence. By adding the possibility of an away step in AFW, or alternatively using the pairwise direction,  zig-zagging can be attenuated. 

The step size $\gamma_t$ can be computed by line-search, i.e.,
\begin{equation} \label{eq:fw:stepsize}
    \gamma_t \in \arg \max_{\gamma \in [0,1]} f(\bx_t + \gamma \bd_t).
\end{equation}
Finally, the Frank-Wolfe gap is used to check if an iterate is (close enough to) a stationary solution.

\begin{mydef} \label{def:fw:gap}
    The Frank-Wolfe gap $g_t$ of $f: \domainD \rightarrow \mathbb{R}$ at iterate $t$ is defined as
    \begin{equation} \label{eq:fw:gap}
        \begin{split}
        g_t = \max\limits_{\bs \in \domainD} \nabla f(\bx_t)^T(\bs - \bx_t)
        \iff
        g_t = \nabla f(\bx_t)^T\bd_t^{FW}.
        \end{split}
    \end{equation}
\end{mydef}
A point $\bx_t$ is stationary if and only if $g_t = 0$, meaning there are no ascent directions. The Frank-Wolfe gap is thus a reasonable measure of nonstationarity and is frequently used as a stopping criterion \cite{lacoste2016}. Specifically, a threshold $\epsilon$ is defined, and if $g_t \leq \epsilon$, then we conclude the $\bx_t$ at the current  iterate $t$ is sufficiently close to a stationary point and we stop the algorithm.

\section{Frank-Wolfe for Dominant Set Clustering}
Here we apply the Frank-Wolfe methods from the previous section to the optimization problem \eqref{eq:stqp1} defined by DSC.

\paragraph{Optimization in Simplex Domain}
Because of the simplex form -- the constraints in \eqref{eq:stqp1} -- the convex combination in \eqref{eq:convexcomb} for $\bx \in \Delta$ can be written as
\begin{equation}
        \bx = \sum_{i=1}^n \lambda_{\be_i} \be_i,
\end{equation}
where $\be_i$ are the standard basis vectors. That is, the $i$-th coefficient corresponds to the $i$-th component of $\bx$, $\lambda_{\be_i} = x_i$. The set of components with nonzero coefficients of $\bx_t$ gives the support, i.e.,
\begin{equation}
    \sigma_t = \seta{i \in V: \xhigh{t}_i > 0}.
\end{equation}

Due to the structure of the simplex $\Delta$, the solution of the optimization \eqref{eq:fwstep1} is
\begin{align} \label{eq:fw:lopt1}
\begin{split}
&\left\{
\begin{array}{ll}
\bs_t &\in \Delta \\
\shigh{t}_i &= 1, \quad \textrm{ where }
i \in \arg \max \limits_{i} \nabla_i f(\bx_t) \\
\shigh{t}_j &= 0, \quad \textrm{ for } j \ne i,
\end{array}
\right.
\end{split}
\end{align}
and the optimization \eqref{eq:fwstep2} is obtained by
\begin{align} \label{eq:fw:lopt2}
\begin{split}
&\left\{
\begin{array}{ll}
\bv_t &\in \Delta \\
\vhigh{t}_i &= 1, \quad \textrm{ where }
i \in \arg \min \limits_{i \in \sigma_t} \nabla_i f(\bx_t) \\
\vhigh{t}_j &= 0, \quad \textrm{ for } j \ne i.
\end{array}
\right.
\end{split}
\end{align}
The maximum and minimum values of the linearization are the largest and smallest components of the gradient, respectively (subject to $i \in \sigma_t$ in the latter case). Note that the gradient is $\nabla f(\bx_t) = 2\bA\bx_t$.

\paragraph{Step Sizes} \label{sec:fw:stepsizes}
We compute the optimal step sizes for FW, PFW, and AFW. Iterate subscripts $t$ are omitted for clarity. We define the step size function as
\begin{align} \label{eq:fw:psi1}
\begin{split}
\psi(\gamma) &= f(\bx + \gamma \bd) \\
&= (\bx + \gamma \bd)^T\bA(\bx + \gamma \bd) \\
&= \bx^T\bA\bx + 2\gamma \bd^T\bA\bx + \gamma^2 \bd^T\bA\bd \\
&= f(\bx) + \gamma \nabla f(\bx_t)^T\bd + \gamma^2 \bd^T\bA\bd
,
\end{split}
\end{align}
for some ascent direction $\bd$. This expression is a single variable second degree polynomial in $\gamma$. The function is concave if the coefficient $\bd^T\bA\bd \leq 0$ -- second derivative test -- and admits a global maximum in that case.

In the following it is assumed that $\bs$ and $\bv$ satisfy \eqref{eq:fw:lopt1} and \eqref{eq:fw:lopt2}, and their nonzero components are $i$ and $j$, respectively.

\noindent\emph{FW direction}: Substitute $\bd^{FW} = \bs - \bx$ into $\bd^T\bA\bd$.

\begin{align} \label{eq:fw:psi2}
    \begin{split}
    \bd^T\bA\bd &= (\bs - \bx)^T\bA(\bs - \bx) \\
    &= \bs^T \bA \bs - 2\bs^T\bA\bx + \bx^T\bA\bx \\
    &= -(2\bs^T\bA\bx - \bx^T\bA\bx) \\
    &= \bx^T\bA\bx - 2\ba_{i*}^T\bx.
    \end{split}
\end{align}

The $i$-th row of $\mathbf A$ is $\ba_{i*}$ and its  $j$-th column is $\ba_{*j}$.  

\noindent\emph{Pairwise direction}: Substitute $\bd^{PFW} = \bs - \bv$ into $\bd^T\bA\bd$.
\begin{align} \label{eq:fw:psi3}
    \begin{split}
    \bd^T\bA\bd &= (\bs - \bv)^T\bA(\bs - \bv) \\
    &= \bs^T \bA \bs - 2\bv^T\bA\bs + \bv^T\bA\bv = -2a_{ij}.
    \end{split}
\end{align}

\noindent\emph{Away direction}: Substitute $\bd^{A} = \bx - \bv$ into $\bd^T\bA\bd$.
\begin{align} \label{eq:fw:psi4}
    \begin{split}
    \bd^T\bA\bd &= (\bx - \bv)^T\bA(\bx - \bv) \\
    &= \bx^T \bA \bx - 2\bv^T\bA\bx + \bv^T\bA\bv \\
    &= \bx^T \bA \bx - 2\ba_{j*}^T\bx
    .
    \end{split}
\end{align}

Recall $\bA$ has nonnegative entries and zeros on the main diagonal. Therefore $\bs^T\bA\bs = 0$ and $\bv^T\bA\bv = 0$. It is immediate that \eqref{eq:fw:psi3} is nonpositive. From $\bx^T\bA\bx \leq \bs^T\bA\bx$ we conclude that \eqref{eq:fw:psi2} is also nonpositive. The corresponding step size functions are therefore always \emph{concave}. We cannot make any conclusion for \eqref{eq:fw:psi4}, and the sign of $\bd^T\bA\bd$  depends on $\bx$.

The derivative of $\psi(\gamma)$ is
\begin{align}
\frac{d \psi}{d \gamma}(\gamma) &= \nabla f(\bx)^T\bd + 2\gamma \bd^T\bA\bd
.
\end{align}

By solving $\frac{d \psi}{d \gamma}(\gamma) = 0$ we obtain
\begin{align} \label{eq:fw:gamma1}
\begin{split}
&\nabla f(\bx)^T\bd + 2\gamma \bd^T\bA\bd = 0 \\
& \iff \\
\gamma^{*} &= -\frac{\nabla f(\bx)^T\bd}{2\bd^T\bA\bd} = -\frac{\bx^T\bA\bd}{\bd^T\bA\bd}
.
\end{split}
\end{align}

Since $\nabla f(\bx)^T\bd \geq 0$, we also conclude here that $\bd^T\bA\bd < 0$ has to hold in order for the step size to make sense.

By substituting the directions and corresponding $\bd^T\bA\bd$ into \eqref{eq:fw:gamma1} we obtain the  optimal step sizes.
\\
\emph{FW direction and \eqref{eq:fw:psi2}}:
\begin{align} \label{eq:fw:gamma2}
\gamma^{FW} &= -\frac{\bx^T\bA\bd}{\bd^T\bA\bd} = \frac{\ba_{i*}^T\bx - \bx^T\bA\bx}{2\ba_{i*}^T\bx - \bx^T\bA\bx}.
\end{align}
\emph{Pairwise direction and \eqref{eq:fw:psi3}}:
\begin{align} \label{eq:fw:gamma3}
\gamma^{PFW} &= -\frac{\bx^T\bA\bd}{\bd^T\bA\bd} = \frac{\ba_{i*}^T\bx - \ba_{j*}^T\bx}{2a_{ij}}.
\end{align}
\emph{Away direction and \eqref{eq:fw:psi4}}:
\begin{align} \label{eq:fw:gamma4}
\gamma^{A} &= -\frac{\bx^T\bA\bd}{\bd^T\bA\bd} = \frac{\bx^T\bA\bx - \ba_{j*}^T\bx}{2\ba_{j*}^T\bx - \bx^T\bA\bx}.
\end{align}

\paragraph{Algorithms}

Here, we describe in detail standard FW (Algorithm \ref{alg:fw:v2}), pairwise FW (Algorithm \ref{alg:fw:v3}), and away-steps FW (Algorithm \ref{alg:fw:v4}) for problem \eqref{eq:stqp1}, following the high-level structure of Algorithm \ref{alg:fw:v1}.
All variants have $\bigO(n)$ per-iteration time complexity, where the linear operations are $\arg\max$, $\arg\min$, and vector addition. The key for this complexity is that we can update the gradient $\nabla f(\bx) = 2\bA\bx$ in linear time. Lemmas \ref{lem:fw:v2}, \ref{lem:fw:pw} and \ref{lem:fw:as} show why this is the case. Recall  the updates in replicator dynamics are quadratic w.r.t. $n$. \footnote{Proofs are provided in the appendix.}


\begin{algorithm}[htb]
    \caption{FW for DSC}
    \label{alg:fw:v2}
    \begin{algorithmic}[1] 
        \Procedure{FW}{$\bA$, $\epsilon$, $T$}
            \State Select $\bx_0 \in \Delta$
            \State $\br_0 := \bA\bx_0$
            \State $f_0 := \br_0^T \bx_0$
            \For{$t = 0, ..., T-1$}
                \State $\bs_t := \be_i$, where $i \in \arg \max \limits_{\ell} \rhigh{t}_\ell$
                \State $g_t := \rhigh{t}_i - f_t$
                \State \textbf{if} $g_t \leq \epsilon$ \textbf{then break}
                \State $\gamma_t := \frac{\rhigh{t}_i - f_t}{2 \rhigh{t}_i - f_t}$
                \State $\bx_{t+1} := (1 - \gamma_t) \bx_t + \gamma_t \bs_t$
                \State $\br_{t+1} := (1 - \gamma_t) \br_t + \gamma_t \ba_{*i}$
                \State $f_{t+1} := (1 - \gamma_t)^2 f_t + 2\gamma_t(1 - \gamma_t)\rhigh{t}_i$
            \EndFor
            \State \textbf{return} $\bx_t$
        \EndProcedure
    \end{algorithmic}
\end{algorithm}

\begin{lem} \label{lem:fw:v2}
    For $\bx_{t+1} = (1-\gamma_t)\bx_t + \gamma_t\bs_t$, lines 11 and 12 in Algorithm \ref{alg:fw:v2} satisfy
    \begin{align*}
        \br_{t+1} = \bA\bx_{t+1},
        f_{t+1} = \bx_{t+1}^T\bA\bx_{t+1}.
    \end{align*}
\end{lem}

\begin{algorithm}[htb]
    \caption{Pairwise FW for DSC}
    \label{alg:fw:v3}
    \begin{algorithmic}[1] 
        \Procedure{PFW}{$\bA$, $\epsilon$, $T$}
            \State Select $\bx_0 \in \Delta$
            \State $\br_0 := \bA\bx_0$
            \State $f_0 := \br_0^T \bx_0$
            \For{$t = 0, ..., T-1$}
                \State $\sigma_t := \seta{i \in V: \xhigh{t}_i > 0}$
                \State $\bs_t := \be_i$, where $i \in \arg \max \limits_{\ell} \rhigh{t}_\ell$
                \State $\bv_t := \be_j$, where $j \in \arg \min \limits_{\ell \in \sigma_t } \rhigh{t}_\ell$
                \State $g_t := \rhigh{t}_i - f_t$
                \State \textbf{if} $g_t \leq \epsilon$ \textbf{then break}
                \State $\gamma_t := \min \left( \xhigh{t}_j, \frac{\rhigh{t}_i - \rhigh{t}_j}{2 a_{ij}} \right)$
                \State $\bx_{t+1} := \bx_t + \gamma_t(\bs_t - \bv_t)$
                \State $\br_{t+1} := \br_t + \gamma_t (\ba_{*i} - \ba_{*j})$
                \State $f_{t+1} := f_t + 2\gamma_t(\rhigh{t}_i - \rhigh{t}_j) - 2\gamma_t^2a_{ij}$
            \EndFor
            \State \textbf{return} $\bx_t$
        \EndProcedure
    \end{algorithmic}
\end{algorithm}

\begin{lem}\label{lem:fw:pw}
    For $\bx_{t+1} = \bx_t +  \gamma_t(\bs_t - \bv_t)$, lines 13 and 14 in Algorithm \ref{alg:fw:v3} satisfy
    \begin{align*}
        \br_{t+1} = \bA\bx_{t+1},
        f_{t+1} = \bx_{t+1}^T\bA\bx_{t+1}.
   \end{align*}
\end{lem}


\begin{algorithm}[htb]
    \caption{Away-steps FW for DSC}
    \label{alg:fw:v4}
    \begin{algorithmic}[1] 
        \Procedure{AFW}{$\bA$, $\epsilon$, $T$}
            \State Select $\bx_0 \in \Delta$
            \State $\br_0 := \bA\bx_0$
            \State $f_0 := \br_0^T \bx_0$
            \For{$t = 0, ..., T-1$}
                \State $\sigma_t := \seta{i \in V: \xhigh{t}_i > 0}$
                \State $\bs_t := \be_i$, where $i \in \arg \max \limits_{\ell} \rhigh{t}_\ell$
                \State $\bv_t := \be_j$, where $j \in \arg \min \limits_{\ell \in \sigma_t } \rhigh{t}_\ell$
                \State $g_t := \rhigh{t}_i - f_t$
                \State \textbf{if} $g_t \leq \epsilon$ \textbf{then break}
                \If{$(\rhigh{t}_i - f_t) \geq (f_t - \rhigh{t}_j)$}
                \Comment{FW direction}
                \State $\gamma_t := \frac{\rhigh{t}_i - f_t}{2\rhigh{t}_i - f_t}$
                \State $\bx_{t+1} := (1 - \gamma_t) \bx_t + \gamma_t \bs_t$
                \State $\br_{t+1} := (1 - \gamma_t) \br_t + \gamma_t \ba_{*i}$
                \State $f_{t+1} := (1 - \gamma_t)^2 f_t + 2\gamma_t(1 - \gamma_t)\rhigh{t}_i$
                \Else
                \Comment{Away direction}
                \State $\gamma_t := \xhigh{t}_j/(1 - \xhigh{t}_j)$
                \If{$(2\rhigh{t}_j - f_t) > 0$}
                \State $\gamma_t \gets \min \left(\gamma_t, \frac{f_t - \rhigh{t}_j}{2\rhigh{t}_j - f_t} \right)$
                \EndIf
                \State $\bx_{t+1} := (1+\gamma_t) \bx_t - \gamma_t \bv_t$
                \State $\br_{t+1} := (1+\gamma_t)\br_t - \gamma_t\ba_{*j}$
                \State $f_{t+1} := (1+\gamma_t)^2f_t - 2\gamma_t(1 + \gamma_t)\rhigh{t}_j$
                \EndIf
            \EndFor
            \State \textbf{return} $\bx_t$
        \EndProcedure
    \end{algorithmic}
\end{algorithm}

Lines 12-15 are identical to the updates in Algorithm \ref{alg:fw:v2}  included in Lemma \ref{lem:fw:v2}. We thus only show the away direction.

\begin{lem}\label{lem:fw:as}
    For $\bx_{t+1} = (1+\gamma_t)\bx_t - \gamma_t\bv_t$, lines 22 and 23 in Algorithm \ref{alg:fw:v4} satisfy
    \begin{align*}
        \br_{t+1} = \bA\bx_{t+1},
        f_{t+1} = \bx_{t+1}^T\bA\bx_{t+1}.
    \end{align*}
\end{lem}

Algorithm \ref{alg:fw:v4} (AFW) is actually equivalent to the \emph{infection and immunization dynamics} (InImDyn) with the pure strategy selection function, introduced in \cite{bulo2011} as an alternative to replicator dynamics. However, InImDyn is derived from the perspective of evolutionary game theory as opposed to Frank-Wolfe. Thus, our framework provides a way to connect replicator dynamics and InImDyn in a principled way. Moreover, it allows us to further analyze this method and study its convergence rate.

\begin{prop} \label{prop:inimdyn}
    Algorithm \ref{alg:fw:v4} (AFW) is equivalent to InImDyn, i.e., Algorithm 1 in \cite{bulo2011}.
\end{prop}

\section{Analysis of Convergence Rates} \label{sec:fw:cr}
\cite{lacoste2016} shows that the Frank-Wolfe gap for standard FW decreases at rate $\bigO(1/\sqrt{t})$ for nonconvex/nonconcave objective functions, where $t$ is the number of iterations. A similar convergence rate is shown in \cite{bomze2019} for nonconvex AFW over the simplex. When the objective function is convex/concave, linear convergence rates for PFW and AFW are shown in \cite{lacoste2015}. The analysis in \cite{ThielCD19} shows linear convergence rate of standard FW for nonconvex but multi-linear functions. We are not aware of any work analyzing the convergence rate in terms of the Frank-Wolfe gap for nonconvex/nonvoncave PFW.

Following the terminology and techniques in \cite{lacoste2016,lacoste2015,bomze2019}, we present a unified and specialized framework to analyze convergence rates for Algorithms \ref{alg:fw:v2}, \ref{alg:fw:v3}, and \ref{alg:fw:v4}. The analysis is split into a number of different cases, where each case handles a unique ascent direction and step size combination. For the step sizes, we consider one case when the optimal step size is used ($\gamma_t < \gamma_{max}$), and a second case when it has been truncated ($\gamma_t = \gamma_{max}$). The former case is referred to as a good step, since in this case we can provide a lower bound on the progress $f(\bx_{t+1}) - f(\bx_t)$ in terms of the Frank-Wolfe gap. The latter case is referred to as a drop step or a swap step. It is called a drop step when the cardinality of the support reduces by one, i.e., $|\sigma_{t+1}| = |\sigma_t|-1$, and it is called a swap step when it remains unchanged, i.e., $|\sigma_{t+1}| = |\sigma_t|$. When $\gamma_t = \gamma_{max}$ we cannot provide a bound on the progress in terms of the Frank-Wolfe gap, and instead we bound the number of drop/swap steps. Furthermore, this case can only happen for PFW and AFW as the step size for FW always satisfies $\gamma_t < \gamma_{max}$. Swap steps can only happen for PFW.

Let
\[
\tilde{g}_t = \min\limits_{0 \leq \ell \leq t} g_\ell, \quad \underline{M} = \min\limits_{i,j: i \ne j} a_{ij}, \quad \overline{M} = \max\limits_{i,j: i \ne j} a_{ij}
\]
be the smallest Frank-Wolfe gap after $t$ iterations, and the smallest and largest off-diagonal elements of $\bA$. Let $I$ be the indexes that take a good step. That is, for $t \in I$ we have $\gamma_t < \gamma_{max}$. Then, we show the following results.

\begin{lem} \label{lem:fw:cr:common}
    The smallest Frank-Wolfe gap for Algorithms \ref{alg:fw:v2}, \ref{alg:fw:v3}, and \ref{alg:fw:v4} satisfy
    \begin{equation} \label{eq:fw:cr:common}
        \tilde{g}_t \leq 2\sqrt{\frac{\beta\left( f(\bx_t) - f(\bx_0) \right)}{|I|}},
    \end{equation}
    where $\beta = 2\overline{M} - \underline{M}$ for FW and AFW, and $\beta = 2\overline{M}$ for PFW.
\end{lem}

\begin{thm} \label{thm:fw:cr:alg2}
    The smallest Frank-Wolfe gap for Algorithm \ref{alg:fw:v2} (FW) satisfies
    \begin{equation} \label{eq:fw:cr:alg2}
        \tilde{g}_t^{FW} \leq 2\sqrt{\frac{(2\overline{M} - \underline{M}) \left( f(\bx_t) - f(\bx_0) \right)}{t}}.
    \end{equation}
\end{thm}

\begin{thm} \label{thm:fw:cr:alg3}
    The smallest Frank-Wolfe gap for Algorithm \ref{alg:fw:v3} (PFW) satisfies
    \begin{equation} \label{eq:fw:cr:alg3}
        \tilde{g}_t^{PFW} \leq 2\sqrt{\frac{6n!\overline{M}\left( f(\bx_t) - f(\bx_0) \right)}{t}}.
    \end{equation}
\end{thm}

\begin{thm} \label{thm:fw:cr:alg4}
    The smallest Frank-Wolfe gap for Algorithm \ref{alg:fw:v4} (AFW) satisfies
    \begin{equation} \label{eq:fw:cr:alg4}
        \tilde{g}_t^{AFW} \leq 2\sqrt{\frac{2(2\overline{M} - \underline{M}) \left( f(\bx_t) - f(\bx_0) \right)}{t+1-|\sigma_0|}}\; .
    \end{equation}
\end{thm}

\noindent From Theorems  \ref{thm:fw:cr:alg2},  \ref{thm:fw:cr:alg3} and \ref{thm:fw:cr:alg4} we conclude Corollary \ref{cor:fw}.

\begin{cor} \label{cor:fw}
    The smallest Frank-Wolfe gap for Algorithms \ref{alg:fw:v2}, \ref{alg:fw:v3}, and \ref{alg:fw:v4} decrease at rate $\bigO(1/\sqrt{t})$.
\end{cor}

\paragraph{Initialization}
The way the algorithms are initialized -- value of $\bx_0$ -- affects the local optima the algorithms converge to.
Let $\bar{\bx}^B = \frac{1}{n} \be$ be the barycenter of the simplex $\Delta$, where $\be^T = (1, 1,..., 1)$. We also define $\bar{\bx}^V$ as
\begin{align} \label{eq:fwinit}
\left\{
\begin{array}{ll}
\bar{\bx}^V &\in \Delta \\
\bar{x}^V_i &= 1, \quad \textrm{ where }
i \in \arg \max \limits_{i} \nabla_i f(\bar{\bx}^B) \\
\bar{x}^V_j &= 0, \quad \textrm{ for } j \ne i.
\end{array}
\right.
\end{align}
Initializing $\bx_0$ with $\bar{\bx}^B$ avoids initial bias to a particular solution as it considers a uniform distribution over the available objects. Since $\nabla f(\bar{\bx}^B) = 2\bA\bar{\bx}^B$, the nonzero component of $\bar{\bx}^V$ corresponds to the row of $\bA$ with largest total sum. Therefore, it is biased to an object that is highly similar to many other objects.

The starting point for replicator dynamics is $\bar{\bx}^{RD} = \bar{\bx}^B$, as used for example in \cite{pavan2003new,pavan2007}. Note that if a component of $\bar{\bx}^{RD}$ starts at zero it will remain at zero for the entire duration of the dynamics according to \eqref{eq:replicatorEq}. Furthermore, $(\bar{\bx}^V)^T\bA\bar{\bx}^V = 0$ since $\bA$ has zeros on the main diagonal, and the denominator in replicator dynamics is then zero for this point. Thus, $\bar{\bx}^V$ is not a viable starting point for replicator dynamics.

The starting point for standard FW is $\bar{\bx}^{FW} = \bar{\bx}^V$, and is found experimentally to work well. As  explained in convergence rate analysis, FW never performs any drop steps since the step size always satisfies $\gamma_t < \gamma_{max}$. Hence, using $\bar{\bx}^B$ as starting point for FW will lead to a solution that has full support -- this is found experimentally to hold true as well. Therefore, with FW, we use only initialization with $\bar{\bx}^V$.
With PFW and AFW, we can use both $\bar{\bx}^B$ and $\bar{\bx}^V$ as starting points. We denote the PFW and AFW variants
by PFW-B, PFW-V, AFW-B, and AFW-V, respectively, to specify the starting point.

\section{Experiments}

In this section, we describe the experimental results of the different optimization methods on synthetic, image segmentation and  20 newsgroup datasets. In addition, we study multi-start  optimization for DSC which can be potentially useful for parallel computations.

\begin{figure}[thb!]
    \centering
    \subfigure[]
    {
    \includegraphics[width=0.22\textwidth]{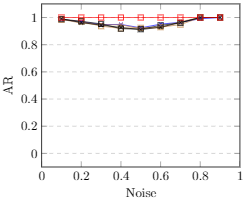}
        \label{fig:synthetic_1_assignrate}
    }
    \subfigure[]
    {
        \includegraphics[width=0.22\textwidth]{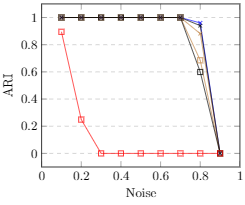}
        \label{fig:synthetic_1_ari}
    }
    \\
    \subfigure[]
    {
    \includegraphics[width=0.22\textwidth]{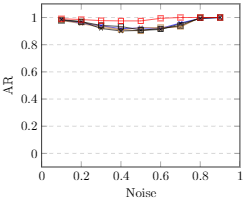}
        \label{fig:synthetic_2_assignrate}
    }
     \subfigure[]
    {
        \includegraphics[width=0.22\textwidth]{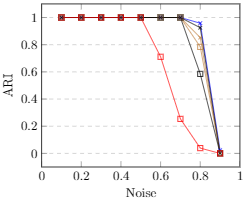}
        \label{fig:synthetic_2_ari}
    }
    \\
    \subfigure[]
    {
        \includegraphics[width=0.22\textwidth]{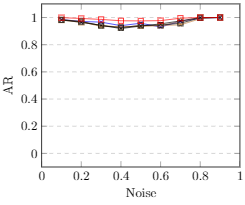}
        \label{fig:synthetic_3_assignrate}
    }
    \subfigure[]
    {
        \includegraphics[width=0.22\textwidth]{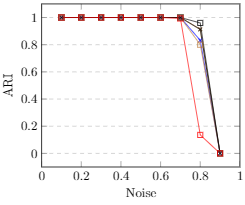}
        \label{fig:synthetic_3_ari}
    }
    \caption{Results on the synthetic dataset for RD (red), FW (blue), PFW (brown), and AFW (black). PFW-B (brown) and AFW-B (black) have squares; PFW-V (brown) and AFW-V (black) have crosses. (a)-(f): $n = 200$. (a, b): $t = 400$, $\delta = 2\cdot10^{-12}$. (c, d): $t = 400$, $\delta = 2\cdot10^{-3}$. (e, f): $t = 4000$, $\delta = 2\cdot10^{-12}$. 
    V-measure demonstrates very similar results to ARI.}
    \label{fig:synth1}
\end{figure}

\subsection{Experimental Setup}

The Frank-Wolfe gap (Definition \ref{def:fw:gap}) and the distance between two consecutive iterates are used as the stopping criterion for the FW variants and replicator dynamics. Specifically, let $\epsilon$ be the threshold, then an algorithm stops if $g_t \leq \epsilon$ or if $\norm{\bx_{t+1} - \bx_t} \leq \epsilon$. In the experiments we set $\epsilon$ to Python's epsilon, $\epsilon \approx 2.2 \cdot 10^{-16}$, and the cutoff parameter $\delta$ to $\delta = 2 \cdot 10^{-12}$.
We denote the true number of clusters in the dataset by $k$ and the maximum number of clusters to extract by $K$.
For a dataset with $n$ objects, the clustering solution is represented by a discrete $n$-dimensional vector $\bc$, i.e., $c_i \in \seta{0, 1, ..., K-1, K}$ for $i = 1,...,n$. If $c_i = c_j$, then objects $i$ and $j$ are in the same cluster. The discrete values $0, 1,..., K-1, K$ are called labels and represent the different clusters. Label 0 is designated to represent `no cluster', i.e., if $c_i = 0$, then object $i$ is unassigned.
We may regularize the pairwise similarities by a shift parameter, as described in detail in \cite{CarlJohnell:Thesis:2020}.

To evaluate the clusterings, we compare the predicted solution and the ground truth w.r.t. Adjusted Rand Index (ARI) \cite{hubert1985} and V-Measure \cite{vmeasure2007}.
The Rand index is the ratio of the object pairs that are either in the same or in different clusters, in both the predicted and ground truth solutions. V-measure is the harmonic mean of  homogeneity and completeness.
We may also report the Assignment Rate (AR), representing the rate of the objects assigned to a valid cluster. As we will discuss, it is common in DSC to apply a postprocessing in order to make AR equal to 1.

\subsection{Experiments on Synthetic Data}

For synthetic experiments, we fix $n=200$ and $K = k = 5$, and assign the objects uniformly to one of the $k$ clusters.

Let $\mu \sim \mathcal{U}(0, 1)$ be uniformly distributed and
\begin{align*}
\left\{
\begin{array}{ll}
z &= 0, \quad \textrm{ with probability } p \\
z &= 1, \quad \textrm{ with probability } 1-p,
\end{array}
\right.
\end{align*}
where $p$ is the noise ratio. The similarity matrix $\bA = (a_{ij})$ is then constructed as follows:
\begin{align*}
\left\{
\begin{array}{lll}
a_{ij} = a_{ji} &= z\mu, \quad &\textrm{ if } i \textrm{ and } j \textrm{ are in the same cluster} \\
a_{ij} &= 0, \quad &\textrm{ otherwise.}
\end{array}
\right.
\end{align*}
For each parameter configuration, we generate a similarity matrix,  perform the clustering five times and then report the average results in Figure \ref{fig:synth1}. We observe that the different FW methods are considerably more robust w.r.t. the noise in  pairwise measurements and yield higher quality results. Also, the performance of FW variants is consistent with  different parameter configurations, whereas RD is more sensitive to the number of iterations $t$ and the cutoff parameter $\delta$.

 \begin{figure*}[thb!]
    \centering
    \subfigure[Original image]
    {
        \includegraphics[width=0.22\textwidth]{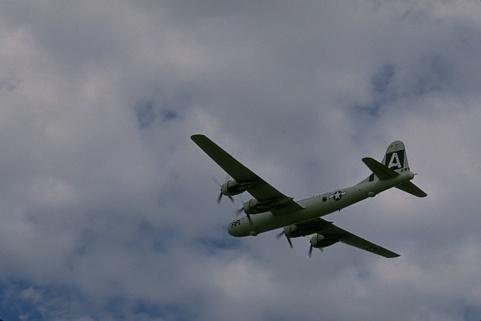}
        \label{fig:airplane}
    }
    \subfigure[RD]
    {
        \includegraphics[width=0.22\textwidth]{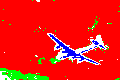}
        \label{fig:airplane_rd2}
    }
    \subfigure[FW, PFW-B, AFW-V, and AFW-B]
    {
        \includegraphics[width=0.22\textwidth]{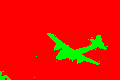}
        \label{fig:airplane_fw2}
    }
    \subfigure[PFW-V]
    {
        \includegraphics[width=0.22\textwidth]{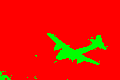}
        \label{fig:airplane_pfwv2}
    }
    \caption{Original image and the segmentation results from different DSC optimization methods.
    }
    \label{fig:airplane_results}
\end{figure*}

\begin{table}[htb!]
    \centering
    \begin{tabular}{r|lllll}
        & $t$ & time & AR & ARI & V-Meas. \\
        \hline
        \multirow{3}{2.6em}{FW}
& 1000 & 0.36s & 0.6325 & 0.4695 & 0.5388 \\
& 4000 & 1.35s & 0.6885 & 0.4593 & 0.5224 \\
& 8000 & 2.41s & 0.6969 & 0.4673 & 0.5325 \\
        \hline
        \multirow{3}{2.6em}{PFW-B}
& 1000 & 0.43s & 0.7429 & 0.1944 & 0.4289 \\
& 4000 & 1.86s & 0.6605 & 0.467 & 0.5327 \\
& 8000 & 2.62s & 0.642 & 0.471 & 0.5335 \\
        \hline
        \multirow{3}{2.6em}{PFW-V}
& 1000 & 0.52s & 0.6471 & 0.5178 & 0.5745 \\
& 4000 & 1.6s & 0.6487 & 0.4565 & 0.5237 \\
& 8000 & 2.47s & 0.642 & 0.471 & 0.5335 \\
        \hline
        \multirow{3}{2.6em}{AFW-B}
& 1000 & 0.35s & 0.8527 & 0.076 & 0.2854 \\
& 4000 & 1.69s & 0.6258 & 0.3887 & 0.5316 \\
& 8000 & 2.93s & 0.6599 & 0.4676 & 0.5328 \\
        \hline
        \multirow{3}{2.6em}{AFW-V}
& 1000 & 0.46s & 0.6415 & 0.5184 & 0.5736 \\
& 4000 & 1.38s & 0.6482 & 0.518 & 0.5754 \\
& 8000 & 2.75s & 0.6476 & 0.4618 & 0.5257 \\
        \hline
        \multirow{3}{2.6em}{RD}
& 1000 & 1.06s & 1.0 & 0.0 & 0.0 \\
& 4000 & 4.56s & 0.9081 & 0.1852 & 0.3003 \\
& 8000 & 11.4s & 0.6997 & 0.4121 & 0.5384 \\
    \end{tabular}
    \caption{The results on  newsgroups1 dataset.}
    \label{tb:ng1}
\end{table}

\begin{table}[htb!]
    \centering
    \begin{tabular}{r|lllll}
        & $t$ & time & AR & ARI & V-Meas. \\
        \hline
        \multirow{3}{2.6em}{FW}
& 1000 & 0.37s & 0.6587 & 0.5594 & 0.5929 \\
& 4000 & 1.38s & 0.6674 & 0.5479 & 0.5866 \\
& 8000 & 2.6s & 0.6679 & 0.5473 & 0.5864 \\
        \hline
        \multirow{3}{2.6em}{PFW-B}
& 1000 & 0.45s & 0.7508 & 0.135 & 0.3555 \\
& 4000 & 1.57s & 0.6172 & 0.6257 & 0.6364 \\
& 8000 & 2.06s & 0.6172 & 0.6257 & 0.6364 \\
        \hline
        \multirow{3}{2.6em}{PFW-V}
& 1000 & 0.59s & 0.6281 & 0.6095 & 0.6241 \\
& 4000 & 1.85s & 0.6172 & 0.6257 & 0.6364 \\
& 8000 & 3.1s & 0.6172 & 0.6257 & 0.6364 \\
        \hline
        \multirow{3}{2.6em}{AFW-B}
& 1000 & 0.41s & 0.8653 & 0.0979 & 0.316 \\
& 4000 & 1.9s & 0.6172 & 0.6257 & 0.6364 \\
& 8000 & 3.39s & 0.6172 & 0.6257 & 0.6364 \\
        \hline
        \multirow{3}{2.6em}{AFW-V}
& 1000 & 0.48s & 0.663 & 0.5548 & 0.5907 \\
& 4000 & 1.75s & 0.6172 & 0.6257 & 0.6364 \\
& 8000 & 3.38s & 0.6172 & 0.6257 & 0.6364 \\
        \hline
        \multirow{3}{2.6em}{RD}
& 1000 & 0.76s & 1.0 & 0.0 & 0.0 \\
& 4000 & 4.67s & 1.0 & 0.1795 & 0.333 \\
& 8000 & 13.52s & 0.7585 & 0.4391 & 0.5161 \\
    \end{tabular}
    \caption{The results on  newsgroups2 dataset.}
    \label{tb:ng2}
\end{table}

\begin{table}[htb!]
    \centering
    \begin{tabular}{r|lllll}
        & $t$ & time & AR & ARI & V-Meas. \\
        \hline
        \multirow{3}{2.6em}{FW}
& 1000 & 0.41s & 0.6756 & 0.5206 & 0.5879 \\
& 4000 & 1.35s & 0.6468 & 0.5309 & 0.5975 \\
& 8000 & 2.63s & 0.6473 & 0.5314 & 0.5978 \\
        \hline
        \multirow{3}{2.6em}{PFW-B}
& 1000 & 0.49s & 0.758 & 0.217 & 0.4617 \\
& 4000 & 1.79s & 0.6468 & 0.5317 & 0.6004 \\
& 8000 & 2.88s & 0.6468 & 0.5317 & 0.6004 \\
        \hline
        \multirow{3}{2.6em}{PFW-V}
& 1000 & 0.56s & 0.6468 & 0.5317 & 0.6004 \\
& 4000 & 1.96s & 0.6468 & 0.5317 & 0.6004 \\
& 8000 & 3.71s & 0.6468 & 0.5317 & 0.6004 \\
        \hline
        \multirow{3}{2.6em}{AFW-B}
& 1000 & 0.37s & 0.8373 & 0.1381 & 0.3594 \\
& 4000 & 1.83s & 0.6462 & 0.5316 & 0.6003 \\
& 8000 & 3.19s & 0.6468 & 0.5317 & 0.6004 \\
        \hline
        \multirow{3}{2.6em}{AFW-V}
& 1000 & 0.49s & 0.6468 & 0.5322 & 0.5993 \\
& 4000 & 1.63s & 0.6468 & 0.5317 & 0.6004 \\
& 8000 & 2.99s & 0.6468 & 0.5317 & 0.6004 \\
        \hline
        \multirow{3}{2.6em}{RD}
& 1000 & 0.86s & 1.0 & 0.0 & 0.0 \\
& 4000 & 4.69s & 0.9089 & 0.2212 & 0.3465 \\
& 8000 & 12.9s & 0.8012 & 0.3526 & 0.4556 \\
    \end{tabular}
    \caption{The results on newsgroups3 dataset.}
    \label{tb:ng3}
\end{table}

\begin{table}[htb!]
    \centering
    \begin{tabular}{r|lllll}
        & $t$ & time & AR & ARI & V-Meas. \\
        \hline
        \multirow{3}{2.6em}{FW}
& 1000 & 0.42s & 0.653 & 0.5097 & 0.5706 \\
& 4000 & 1.38s & 0.6169 & 0.4672 & 0.5437 \\
& 8000 & 2.6s & 0.7002 & 0.5014 & 0.5514 \\
        \hline
        \multirow{3}{2.6em}{PFW-B}
& 1000 & 0.43s & 0.8092 & 0.2247 & 0.4483 \\
& 4000 & 1.82s & 0.6697 & 0.6211 & 0.6484 \\
& 8000 & 3.04s & 0.6697 & 0.6211 & 0.6484 \\
        \hline
        \multirow{3}{2.6em}{PFW-V}
& 1000 & 0.58s & 0.6591 & 0.6446 & 0.6717 \\
& 4000 & 2.02s & 0.6565 & 0.6462 & 0.675 \\
& 8000 & 2.74s & 0.6565 & 0.6462 & 0.675 \\
        \hline
        \multirow{3}{2.6em}{AFW-B}
& 1000 & 0.35s & 0.9041 & 0.1109 & 0.3361 \\
& 4000 & 1.87s & 0.6687 & 0.6191 & 0.6463 \\
& 8000 & 3.55s & 0.6697 & 0.6211 & 0.6484 \\
        \hline
        \multirow{3}{2.6em}{AFW-V}
& 1000 & 0.5s & 0.6525 & 0.5071 & 0.5651 \\
& 4000 & 1.84s & 0.6565 & 0.6462 & 0.675 \\
& 8000 & 3.6s & 0.6565 & 0.6462 & 0.675 \\
        \hline
        \multirow{3}{2.6em}{RD}
& 1000 & 0.93s & 1.0 & 0.0 & 0.0 \\
& 4000 & 5.46s & 1.0 & 0.3197 & 0.4112 \\
& 8000 & 14.52s & 0.8559 & 0.4528 & 0.5328 \\
    \end{tabular}
    \caption{The results on newsgroups4 dataset.}
    \label{tb:ng4}
\end{table}

\begin{table*}[!ht]
    \begin{center}
        \begin{tabular}{ l || l || ll || ll || ll || ll }
         &  & \multicolumn{2}{c||}{newsgroups1}&\multicolumn{2}{c||}{newsgroups2}&\multicolumn{2}{c||}{newsgroups3}&\multicolumn{2}{c}{newsgroups4}\\
        \hline
        Method & t & ARI & V-Meas. & ARI & V-Meas. & ARI & V-Meas. & ARI & V-Meas.  \\
        \hline
        \multirow{3}{4em}{FW}
& 1000 & 0.4068 & 0.4969 & 0.5158 & 0.5733 & 0.5751 & 0.595 & 0.4663 & 0.5252 \\
& 4000 & 0.4639 & 0.5225 & 0.5084 & 0.5699 & 0.572 & 0.5962 & 0.4409 & 0.5084 \\
& 8000 & 0.4766 & 0.5351 & 0.5084 & 0.5699 & 0.5729 & 0.5972 & 0.4973 & 0.5396  \\
        \hline
        \multirow{3}{4em}{PFW-B}
& 1000 & 0.2063 & 0.3919 & 0.178 & 0.3814 & 0.2764 & 0.4859 & 0.288 & 0.4878 \\
& 4000 & 0.4623 & 0.5324 & 0.5332 & 0.5834 & 0.5734 & 0.5992 & 0.587 & 0.6094 \\
& 8000 & 0.4356 & 0.5219 & 0.5332 & 0.5834 & 0.5734 & 0.5992 & 0.587 & 0.6094 \\
        \hline
        \multirow{3}{4em}{PFW-V}
& 1000 & 0.5091 & 0.5763 & 0.5331 & 0.5824 & 0.5734 & 0.5992 & 0.605 & 0.6226 \\
& 4000 & 0.4298 & 0.5149 & 0.5332 & 0.5834 & 0.5734 & 0.5992 & 0.6072 & 0.6268 \\
& 8000 & 0.4356 & 0.5219 & 0.5332 & 0.5834 & 0.5734 & 0.5992 & 0.6072 & 0.6268 \\
        \hline
        \multirow{3}{4em}{AFW-B}
& 1000 & 0.0966 & 0.2751 & 0.131 & 0.344   & 0.1782 & 0.3967 & 0.1313 & 0.3577 \\
& 4000 & 0.3162 & 0.4806 & 0.5332 & 0.5834 & 0.5734 & 0.5992 & 0.588 & 0.6097 \\
& 8000 & 0.4615 & 0.5319 & 0.5332 & 0.5834 & 0.5734 & 0.5992 & 0.587 & 0.6094 \\
        \hline
        \multirow{3}{4em}{AFW-V}
& 1000 & 0.5066 & 0.5744 & 0.5099 & 0.5699 & 0.5741 & 0.5983 & 0.4592 & 0.5183 \\
& 4000 & 0.5047 & 0.5719 & 0.5332 & 0.5834 & 0.5734 & 0.5992 & 0.6072 & 0.6268 \\
& 8000 & 0.4308 & 0.5148 & 0.5332 & 0.5834 & 0.5734 & 0.5992 & 0.6072 & 0.6268 \\
        \hline
        \multirow{3}{4em}{RD}
& 1000 & 0.0 & 0.0 & 0.0 & 0.0 & 0.0 & 0.0 & 0.0 & 0.0  \\
& 4000 & 0.1892 & 0.3042 & 0.1795 & 0.333 & 0.2394 & 0.3575 & 0.3197 & 0.4112 \\
& 8000 & 0.3659 & 0.4937 & 0.4123 & 0.5065 & 0.4227 & 0.4973 & 0.4858 & 0.5556 \\
        \end{tabular}
    \end{center}
    \caption{Result of different methods on 20 newsgroup datasets after post assignment of the unassigned documents. The FW variants especially PFW-V and AFW-V yield the best and computationally the most efficient results even with $t = 1000$.
    }
    \label{tb:ngpost}
\end{table*}

\subsection{Image Segmentation}
Next, we study segmentation of colored images in HSV space. We define the feature vector $\bmf(i) = \lbrack v, vs \sin(h), vs \cos(h) \rbrack^T$ as in \cite{pavan2007}, where $h$, $s$, and $v$ are the HSV values of pixel $i$. The similarity matrix $\bA$ is then defined as follows. (i) Compute $\norm{\bmf(i) - \bmf(j)}$, for every pair of pixels $i$ and $j$ to obtain $\bD^{L2}$. (ii) Compute the minimax (path-based) distances \cite{fischer2003,Chehreghani20minimax,chehreghani2017} from $\bD^{L2}$ to obtain $\bD^{P}$. (iii) Finally, $\bA = \max (\bD^{P}) - \bD^{P}$, where max is over the elements in $\bD^{P}$ as used in \cite{chehreghani2016,ChehreghaniC20}.
Figure \ref{fig:airplane_results} shows the segmentation results of the airplane image in Figure \ref{fig:airplane}. The image has the dimensions $120 \times 80$, which leads to a clustering problem with $n = 120 \times 80 = 9600$ objects. We run the FW variants for $t=10000$ and RD for $t=250$ iterations. Due to the linear versus quadratic per-iteration time complexity of the FW variants and RD, we are able to run FW for many more iterations. This allows us to have more flexibility in tuning the parameters and thus obtain more robust results.
According to the results in Figure \ref{fig:airplane_results}, the FW variants visually yield more meaningful and consistent results compared to RD, and separate better the airplane from the background.

\subsection{Experiments on 20 Newsgroups Data}
We study the clustering of different subsets of 20 newsgroups data collection. The collection consists of $18000$ documents in 20 categories split into training and test subsets. We use four datasets with documents from randomly selected categories from the test subset.
(i) \emph{newsgroups1}: the set of documents in the categories soc.religion.christian, comp.os.ms-windows.misc, talk.politics.guns, alt.atheism, talk.politics.misc.
(ii) \emph{newsgroups2}: the set of documents in the categories comp.windows.x, sci.med, rec.autos, sci.crypt, talk.religion.misc.
(iii) \emph{newsgroups3}: the set of documents in the categories misc.forsale, comp.sys.mac.hardware, talk.politics.mideast, sci.electronics, rec.motorcycles.
(iv) \emph{newsgroups4}: the set of documents in the categories  comp.graphics, rec.sport.hockey, sci.space, \\ rec.sport.baseball, comp.sys.ibm.pc.hardware.

Each dataset has $k=5$ true clusters and $1700 \leq n \leq 2000$ documents, where we use $K=5$ for peeling off the computed clusters.
We obtain the tf-idf (term-frequency times inverse document-frequency) vector for each document and then apply PCA to reduce the dimensionality to 20. We obtain the similarity matrix $\bA$ using the cosine similarity between the PCA vectors and then shift the off-diagonal elements by $1$ to ensure nonnegative entries.
We set the regularization parameter to $\alpha = 15$, that seems a reasonable choice for different methods. Using smaller or larger $\alpha$ results in too small clusters or too slow convergence, in particular for replicator dynamics (RD).

Tables \ref{tb:ng1}, \ref{tb:ng2}, \ref{tb:ng3}, and \ref{tb:ng4} show the results for the different datasets. We observe that different variants of FW yield significantly better results compared to replicator dynamics (RD), w.r.t. both ARI and V-Measure. In particular, PFW-V and AFW-V are computationally efficient and perform very well even with $t=1000$. On the other hand, these methods are more robust w.r.t. different parameter settings.
Since all the objects in the ground truth solutions are assigned to a cluster, the assignment rate (AR) indicates the ratio of the objects assigned (correctly or incorrectly) to a cluster during the clustering. A high AR does not necessarily indicate a good clustering solution, rather it may imply slower convergence (as happens for RD).  High AR and low ARI/V-measure means assignment of many objects to wrong clusters. This is what happens for RD with $t=1000$.
As discussed in \cite{pavan2007}, it is common for DSC to  perform a post processing to assign each unassigned object to the cluster which it has the highest average similarity with. Specifically, let $C_0 \subseteq V$ contain the unassigned objects and $C_i \subseteq V$, $1 \leq i \leq K$, be the predicted clusters. Object $j \in C_0$ is then assigned to cluster $C_i$ that satisfies
\begin{equation}
i \in \arg \max\limits_{\ell \geq 1} \frac{1}{|C_\ell|} \sum_{p \in C_\ell} \mathbf A_{jp}.
\end{equation}

Table \ref{tb:ngpost} shows the performance of different methods after assigning all the documents to valid clusters, i.e., when AR is always 1.
We observe that ARI and V-measure are usually consistent for pre and post assignment settings. In both cases the FW variants (especially PFW-V and AFW-V) yield the best and computationally the most efficient results. Consistent to the previous results, PFW-V and AFW-V yield high scores already with $t = 1000$. These results are consistent with the results on synthetic and image datasets.

\begin{figure*}[thb]
    \centering
    \subfigure[No noise]
    {
        \includegraphics[width=0.22\textwidth]{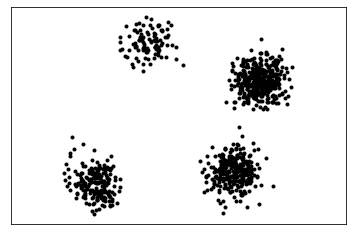}
        \label{fig:ms1:normal1}
    }
    \subfigure[AR 0.94, ARI 1.0, and V-measure 1.0]
    {
        \includegraphics[width=0.22\textwidth]{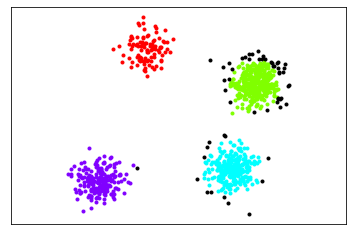}
        \label{fig:ms1:normal2}
    }
    \subfigure[Noise $p=0.4$]
    {
        \includegraphics[width=0.22\textwidth]{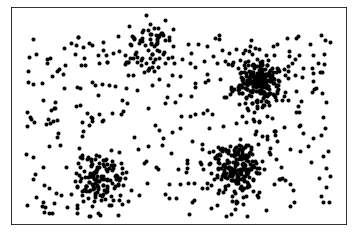}
        \label{fig:ms1:noise1}
    }
    \subfigure[AR 0.74, ARI 1.0, and V-measure 1.0]
    {
        \includegraphics[width=0.22\textwidth]{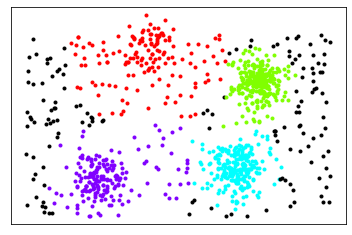}
        \label{fig:ms1:noise2}
    }
    \caption{Two  example datasets used for multi-start study. (b) and (d) show the FW clustering results; PFW and AFW produce similar results.}
    \label{fig:ms1}
\end{figure*}

\begin{figure*}[ht]
    \centering
    \subfigure[]
    {
        \includegraphics[width=0.22\textwidth]{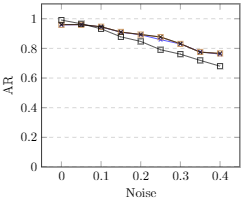}
        \label{fig:multistart_uni_assignrate}
    }
    \subfigure[]
    {
        \includegraphics[width=0.22\textwidth]{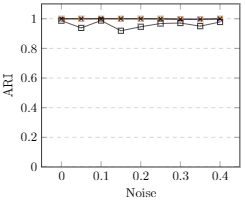}
        \label{fig:multistart_uni_ari}
    }
    \subfigure[]
    {
        \includegraphics[width=0.22\textwidth]{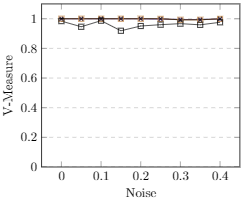}
        \label{fig:multistart_uni_vmeasure}
    }
    \subfigure[]
    {
        \includegraphics[width=0.22\textwidth]{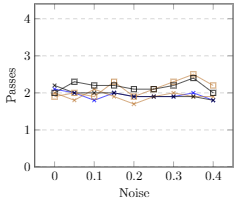}
        \label{fig:multistart_uni_passes}
    }
    \\
    \subfigure[]
    {
        \includegraphics[width=0.22\textwidth]{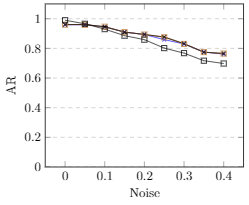}
        \label{fig:multistart_dpp_assignrate}
    }
    \subfigure[]
    {
        \includegraphics[width=0.22\textwidth]{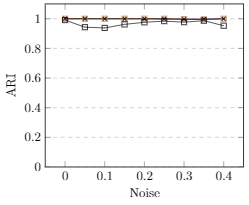}
        \label{fig:multistart_dpp_ari}
    }
    \subfigure[]
    {
        \includegraphics[width=0.22\textwidth]{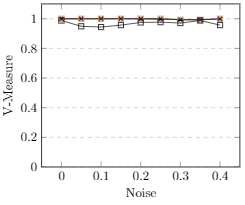}
        \label{fig:multistart_dpp_vmeasure}
    }
    \subfigure[]
    {
        \includegraphics[width=0.22\textwidth]{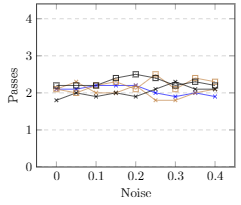}
        \label{fig:multistart_dpp_passes}
    }
    \caption{Results of multi-start paradigm with FW (blue) , PFW (brown), and AFW (black) where  PFW-B (brown) and AFW-B (black) are marked by squares; and  PFW-V (brown) and AFW-V (black) are marked by crosses. The first row corresponds to UNI and the second row corresponds to DPP sampling.
    All optimization and sampling methods require only about two passes to compute the clusters.
    }
    \label{fig:ms2}
\end{figure*}

\subsection{Multi-Start Dominant Set Clustering}
Finally, as a side study, we study a combination of multi-start Dominant Set Clustering with the peeling off strategy. For this, we perform the following procedure.
    1. Sample a subset of objects, and use them to construct a number of starting points for the same similarity matrix.
    2.  Run an optimization method for each starting point.
    3. Identify the nonoverlapping clusters from the solutions and remove (peel off) the corresponding objects from the similarity matrix.
    4. Repeat until no objects are left or a sufficient number of clusters have been found.

This scenario can be potentially useful in particular when multiple processors can perform clustering in parallel.
However, if all the different starting points converge to the same cluster, then there would be no computational benefit.
Thus, here we investigate such a possibility for our optimization framework. For this, we consider the number of passes through the entire data, where a pass is defined as one complete run of the aforementioned steps.
After the solutions from a pass are computed, they are sorted based on the function value $f(\bx)$. The sorted solutions are then permuted in a decreasing order, and if the support of the current solution overlaps more than $10\%$ with the support of the other (previous) solutions, it is discarded. Each pass will therefore yield at least one new cluster. With $K$ the maximum number of clusters to extract, there will be at most $K$ passes. Thus in order for the method to be useful and effective, less than $K$ passes should be performed.

Figure \ref{fig:ms1} shows the form of the datasets used in this study. Each cluster corresponds to a two dimensional Gaussian distribution with a fixed mean and an identity co-variance matrix (see Figure \ref{fig:ms1:normal1}). We fix $n=1000$ and $K=k=4$, and use the parameter $p$ to control the noise ratio. Set $n_1 = pn$ and $n_2 = n - n_1$. A dataset is then generated by sampling $n_1$ objects from a uniform distribution (background noise in Figure \ref{fig:ms1:noise1}), $0.1\cdot n_2$, $0.2\cdot n_2$, $0.3\cdot n_2$, and $0.4\cdot n_2$ objects from the respective Gaussians.

Let $\bD$ be the matrix with pairwise Euclidean distances between all objects in the dataset. The similarity matrix is then defined as $\bA = \max(\bD) - \bD$, similar to the image segmentation study but with a different base distance measure. The regularization parameter is set to $\alpha = 15$.
To determine the starting points we sample 4 components from $\seta{1, ..., n}$, denoted by $i_1, i_2, i_3, i_4$. The number 4 matches the number of CPUs in our machine. For a given component $i \in \seta{i_1, i_2, i_3, i_4}$, we define the starting points as

\begin{align*}
\left\{
\begin{array}{ll}
\bar{x}^V_i &= 1 \\
\bar{x}^V_j &= 0, \quad \textrm{ for } j \ne i
\end{array}
\right.
\end{align*}
and
\begin{align*}
\left\{
\begin{array}{ll}
\bar{x}^B_i &= 0.5 \\
\bar{x}^B_j &= 0.5/(n-1), \quad \textrm{ for } j \ne i.
\end{array}
\right.
\end{align*}
FW uses only $\bar{\bx}^V$ while PFW and AFW use both $\bar{\bx}^V$ and $\bar{\bx}^B$.
To sample the components, we use uniform sampling and Determinantal Point Processes \cite{kulesza2012}, denoted as UNI and DPP, respectively.

Figure \ref{fig:ms2} illustrates the results for the different sampling methods and starting objects. For a given dataset, sampling method, and optimization method, we generate starting objects and run the experiments 10 times and report the average results. Each method is run for $t=1000$ iterations. For this type of dataset we do not observe any significant difference between FW, PFW, or AFW when using either DPP or UNI. It seems that AFW with $\bar{\bx}^B$ as starting object performs slightly worse.

However, we observe that all the sampling and optimization methods require only two passes, whereas we have $K=4$. This observation implies that the multi-start paradigm is potentially useful for computing the clusters in parallel with the different FW variants. We note that the peeling of strategy in Dominant Set Clustering is inherently sequential, thus such an achievement can be potentially very useful for parallelization.

\section{Conclusion}
We developed a unified and computationally efficient framework to employ the different variants of Frank-Wolfe methods for Dominant Set Clustering. In particular, replicator dynamics was replaced with standard, pairwise, and away-steps Frank-Wolfe algorithms when optimizing the quadratic problem defined by DSC. We provided a specialized analysis of the algorithms' convergence rates, and demonstrated the effectiveness of the framework via experimental studies on synthetic, image and text datasets. We additionally studied aspects such as multi-start Dominant Set Clustering.
Our framework is generic enough to be investigated as an alternative for replicator dynamics in many other problems.

\section*{Acknowledgement}
The work of Morteza Haghir Chehreghani was partially supported by the Wallenberg AI, Autonomous Systems and Software Program (WASP) funded by the Knut and Alice Wallenberg Foundation.

\bibliographystyle{plain}
\balance
\bibliography{main}

\section*{Appendix  - Proofs}

\subsection*{Proof of Lemma 1} 

\begin{proof}
    By definition (lines 3 and 4 in Algorithm \ref{alg:fw:v2}), $\br_0 = \bA\bx_0$ and $f_0 = \bx_0^T\bA\bx_0$.
    Let $\bx = \bx_t$, $\bs = \bs_t$, and $\gamma = \gamma_t$. Assume $\br_t = \bA\bx$ and $f_t = \bx^T\bA\bx$ holds. Expand the definition of $\bA\bx_{t+1}$ and proceed by induction.
    \begin{align*}
    \bA \bx_{t+1} &= \bA((1-\gamma)\bx + \gamma\bs) \\
    &= (1 - \gamma) \bA\bx + \gamma \bA \bs \\
    &= (1 - \gamma) \br_t + \gamma \ba_{*i} \\
    &= \br_{t+1}, \\
    \bx_{t+1}^T \bA \bx_{t+1} &= ((1 - \gamma)\bx + \gamma\bs)^T \bA ((1 - \gamma)\bx + \gamma \bs) \\
    &= (1 - \gamma)^2 \bx^T\bA\bx + 2\gamma (1 - \gamma) \bs^T\bA\bx + \gamma^2 \bs^T \bA \bs \\
    &= (1 - \gamma)^2 \bx^T\bA\bx + 2\gamma (1 - \gamma) \bs^T\bA\bx \\
    &= (1 - \gamma)^2 f_t + 2\gamma(1 - \gamma) \rhigh{t}_i \\
    &= f_{t+1}.
    \end{align*}
    Note $\bs^T\bA\bs = 0$ from the definition of $\bs$ and $\bA$.
\end{proof}

\subsection*{Proof of Lemma 2} 

\begin{proof}
    Proceed as in proof of Lemma \ref{lem:fw:v2}. Let $\bx = \bx_t$, $\bs = \bs_t$, $\bv = \bv_t$, and $\gamma = \gamma_t$.
    \begin{align*}
    \bA \bx_{t+1} &= \bA(\bx + \gamma(\bs - \bv)) \\
    &= \bA\bx + \gamma(\bA\bs - \bA\bv) \\
    &= \br_{t} + \gamma (\ba_{*i} - \ba_{*j}) \\
    &= \br_{t+1}, \\
    \bx_{t+1}^T \bA \bx_{t+1} &= (\bx + \gamma(\bs - \bv))^T \bA (\bx + \gamma(\bs - \bv)) \\
    &= \bx^T\bA\bx + 2\gamma(\bs - \bv)^T\bA\bx\\
    & \quad + \gamma^2(\bs - \bv)^T\bA(\bs - \bv) \\
    &= \bx^T\bA\bx + 2\gamma(\bs^T\bA\bx - \bv^T\bA\bx) - 2\gamma^2 a_{ij} \\
    &= f_t + 2\gamma(\rhigh{t}_i - \rhigh{t}_j) - 2\gamma^2 a_{ij} \\
    &= f_{t+1}.
    \end{align*}
\end{proof}

\subsection*{Proof of Lemma 3}
\begin{proof}
    Proceed as in proof of Lemma \ref{lem:fw:v2}. Let $\bx = \bx_t$, $\bv = \bv_t$, and $\gamma = \gamma_t$.
    \begin{align*}
        \bA\bx_{t+1} &= \bA((1 + \gamma)\bx - \gamma\bv) \\
        &= (1 + \gamma) \bA\bx - \gamma\bA\bv \\
        &= (1 + \gamma) \br_t - \gamma\ba_{*j} \\
        &= \br_{t+1}, \\
        \bx_{t+1}^T \bA \bx_{t+1} &= ((1 + \gamma)\bx - \gamma\bv)^T \bA ((1 + \gamma)\bx - \gamma \bv) \\
        &= (1 + \gamma)^2 \bx^T\bA\bx \\
        & \quad - 2\gamma (1 + \gamma) \bv^T\bA\bx + \gamma^2 \bv^T \bA \bv \\
        &= (1 + \gamma)^2 \bx^T\bA\bx - 2\gamma (1 + \gamma) \bv^T\bA\bx \\
        &= (1 + \gamma)^2 f_t - 2\gamma(1 + \gamma) \rhigh{t}_j \\
        &= f_{t+1}.
    \end{align*}
\end{proof}

\subsection*{Proof of Lemma 4} 

\begin{proof}
    Let $\by = \bx_t + \gamma_t\bd_t$, for some ascent direction $\bd_t$, $\br(\bx) = \bA\bx$, and $f(\bx) = \bx^T\bA\bx$. From \eqref{eq:fw:psi1} we have
\begin{align*}
    f(\by) &= f(\bx_t) + 2\gamma_t\br(\bx_t)^T\bd_t + \gamma_t^2\bd_t^T\bA\bd_t.
\end{align*}
Using
\begin{align*}
    \gamma_t = -\frac{(\bx_t)^T\bA\bd_t}{\bd_t\bA\bd_t} = -\frac{\br(\bx_t)^T\bd_t}{\bd_t^T\bA\bd_t}
\end{align*}
from \eqref{eq:fw:gamma1}, we get
\begin{align} \label{eq:fw:cr:iterate2}
    f(\by) &= f(\bx_t) - 2\frac{(\br(\bx_t)^T\bd_t)^2}{\bd_t^T\bA\bd_t} + \frac{(\br(\bx_t)^T\bd_t)^2}{\bd_t^T\bA\bd_t} \nonumber\\
    &= f(\bx_t) - \frac{(\br(\bx_t)^T\bd_t)^2}{\bd_t^T\bA\bd_t} \nonumber\\
    & \iff \\
    & \hspace{-7mm} (\br(\bx_t)^T\bd_t)^2 = -\bd_t^T\bA\bd_t \left( f(\by) - f(\bx_t) \right) \nonumber
    .
\end{align}

    Let $\bs_t$ satisfy \eqref{eq:fw:lopt1} and $\bv_t$ satisfy \eqref{eq:fw:lopt2}. Denote their nonzero components by $i$ and $j$, respectively. Let $h_t = f(\bx_{t+1}) - f(\bx_t)$ and $g_t = 2(r(\bx_t)_i - f(\bx_t))$.

    We consider the FW, away, and pairwise directions $\bd_t$ and corresponding step sizes satisfying $\gamma_t < \gamma_{max}$. Note that $f(\by) = f(\bx_{t+1})$ holds in \eqref{eq:fw:cr:iterate2} for such directions and step sizes.

    \emph{FW direction}: Substitute $\bd_t = \bs_t - \bx_t$ and \eqref{eq:fw:psi2} into \eqref{eq:fw:cr:iterate2}.
    \begin{align*}
        \begin{split}
        &(r(\bx_t)_i - f(\bx_t))^2 = (2r(\bx_t)_i - f(\bx_t)) h_t\\
        &\implies
        g_t^2 \leq 4(2\overline{M} - \underline{M})h_t.
        \end{split}
    \end{align*}

    \emph{Away direction}: For this direction with $\gamma_t < \gamma_{max}$ we have
    \[
       r(\bx_t)_i - f(\bx_t) < f(\bx_t) - r(\bx_t)_j,
    \]
    from line 11 in Algorithm \ref{alg:fw:v4}.
    Substitute $\bd_t = \bx_t - \bv_t$ and \eqref{eq:fw:psi4} into \eqref{eq:fw:cr:iterate2}.
    \begin{align*}
        \begin{split}
        &(f(\bx_t) - r(\bx_t)_j)^2 = (2r(\bx_t)_j - f(\bx_t)) h_t \\
        &\implies
        g_t^2 \leq 4(2\overline{M} - \underline{M})h_t.
        \end{split}
    \end{align*}

    \emph{Pairwise direction}: Substitute $\bd_t = \bs_t - \bv_t$ and \eqref{eq:fw:psi3} into \eqref{eq:fw:cr:iterate2}.
    \begin{align*}
        \begin{split}
            &(r(\bx_t)_i - r(\bx_t)_j)^2 = 2a_{ij}h_t \\
            &\implies
            g_t^2 \leq 8\overline{M}h_t.
        \end{split}
    \end{align*}

    Using previously defined $I$ and $\beta$ in section Analysis of Convergence Rates, we get
    \begin{align*}
        4\beta \left(f(\bx_t) - f(\bx_0) \right) &= 4\beta \sum_{\ell=0}^{t-1} h_\ell
        \geq 4\beta \sum_{\ell \in I} h_\ell \\ & \geq \sum_{\ell \in I}g_t^2  \geq |I|\tilde{g}_t^2 \\
        &\implies \\
        &\tilde{g}_t^2 \leq \frac{4\beta \left(f(\bx_{t}) - f(\bx_{0}) \right)}{|I|}
        &\\
        &\iff
        \\
        &\tilde{g}_t \leq 2\sqrt{\frac{\beta \left(f (\bx_t) - f(\bx_{0}) \right)}{|I|}}
        ,
    \end{align*}
    for either direction $\bd_t$.
\end{proof}

\subsection*{Proof of Theorem 1} 

\begin{proof}
    Since standard FW only takes good steps we have $|I| = t$. The result follows from Lemma \ref{lem:fw:cr:common}.
\end{proof}

\subsection*{Proof of Theorem 2} 

\begin{proof}
    When $\gamma_t = \gamma_{max}$ we either have $|\sigma_{t+1}| = |\sigma_t|-1$ or $|\sigma_{t+1}| = |\sigma_t|$, called drop and swap step, respectively. We need to upper bound the number of these steps in order to get a lower bound for $|I|$.

    The following reasoning is from the analysis of PFW with convex objective function in \cite{lacoste2015}.

    Let $n$ be the dimension of $\bx_t$, $m = |\sigma_t|$, and $\bd_t = \bs_t - \bv_t$. Since we are performing line search, we always have $f(\bx_{\ell}) < f(\bx_t)$ for all $\ell < t$ that are nonstationary. This means the sequence $\bx_0, ..., \bx_t$ will not have any duplicates. The set of component values does not change when we perform a swap step:
    \[
        \seta{\xhigh{t}_\ell: \ell = 1,...,n} \cap \seta{\xhigh{t+1}_\ell: \ell = 1,...,n} = \emptyset.
    \]
    That is, the components are simply permuted after a swap step. The number of possible unique permutations is $\kappa = n!/(n-m)!$. After we have performed $\kappa$ swap steps, a drop step can be taken which will change the component values. Thus in the worst case, $\kappa$ swap steps followed by a drop step will be performed until $m=1$ before a good step is taken. The number of swap/drop steps between two good steps is then bounded by
    \[
        \sum_{\ell=1}^m \frac{n!}{(n-\ell)!} \leq n!\sum_{\ell=0}^{\infty} \frac{1}{\ell!} = n!e \leq 3n!.
    \]
    Result \eqref{eq:fw:cr:alg3} follows from Lemma \ref{lem:fw:cr:common} and
    \[
        |I| \geq \frac{t}{3n!} \; .
    \]
\end{proof}

\subsection*{Proof of Theorem 3}

\begin{proof}
    When $\gamma_t = \gamma_{max}$, $\bd_t$ must be the away direction. In this case the support is reduced by one, i.e. $|\sigma_{t+1}| = |\sigma_t|-1$. Denote these indexes by $D$. Let $I_A \subseteq I$ be the indexes that adds to the support, i.e. $|\sigma_{t+1}| > |\sigma_t|$ for $t \in I_A$. Similar as before, we need to upper bound $|D|$ in order to get a lower bound for $|I|$.

    We have $|I_A| + |D| \leq t$ and $|\sigma_t| = |\sigma_0| + |I_A| - |D]$. Combining the inequalities we get
    \begin{align*}
        &1 \leq |\sigma_t| \leq |\sigma_0| + t - 2|D|  \\
        \implies&
        |D| \leq \frac{|\sigma_0| - 1 + t}{2} .
    \end{align*}
    Result \eqref{eq:fw:cr:alg4} then follows from Lemma \ref{lem:fw:cr:common} and
    \begin{align*}
        |I| = t - |D| \geq t - \frac{(|\sigma_0| - 1 + t)}{2} = \frac{t + 1 - |\sigma_0|}{2} \; .
    \end{align*}
\end{proof}

\end{document}